\documentclass[letterpaper, 10 pt, conference]{ieeeconf} 
\IEEEoverridecommandlockouts
\title{\bf Enhanced-FQL($\lambda$), an Efficient and Interpretable RL with novel Fuzzy Eligibility Traces and Segmented Experience Replay}

\author{{Mohsen Jalaeian-Farimani, Xiong Xiong, and Luca Bascetta}
\thanks{Authors are with the Dipartimento di Elettronica, informazione e Bioingegneria (DEIB), Politecnico di Milano, 20133 Milan, Italy. E-mails: \{mohsen.jalaeian, xiong.xiong, and  luca.bascetta\}@polimi.it}
}

\usepackage{amsmath,amssymb,amsfonts}
\usepackage{algorithm}
\usepackage{graphicx}  %
\usepackage{xcolor}
\usepackage[noend]{algpseudocode}
\newtheorem{theorem}{Theorem}

\newtheorem{remark}{Remark}
\newtheorem{definition}{Definition}
\usepackage{hyperref}
\begin{document}
\maketitle
\thispagestyle{empty}
\pagestyle{empty}

\begin{abstract}
This paper introduces a fuzzy reinforcement learning framework, Enhanced-FQL($\lambda$), that integrates novel Fuzzified Eligibility Traces (FET) and Segmented Experience Replay (SER) into fuzzy Q-learning with the Fuzzified Bellman Equation (FBE) for continuous control. The proposed approach employs an interpretable fuzzy rule base instead of complex neural architectures, while maintaining competitive performance through two key innovations: a fuzzified Bellman equation with eligibility traces for stable multi-step credit assignment, and a memory-efficient segment-based experience replay mechanism for enhanced sample efficiency. Theoretical analysis proves the proposed method convergence under standard assumptions. On the Cart--Pole benchmark, Enhanced-FQL($\lambda$) improves sample efficiency and reduces variance relative to $n$-step fuzzy TD and fuzzy SARSA($\lambda$), while remaining competitive with the tested DDPG baseline. These results support the proposed framework as an interpretable and computationally compact alternative for moderate-scale continuous control problems.
\end{abstract}


\section{INTRODUCTION}
Designing intelligent agents that can autonomously learn and adapt in high-dimensional, dynamic environments, characterized by pervasive uncertainty and the need for online decision-making, remains a significant challenge in artificial intelligence. To address this core problem, reinforcement learning (RL) has emerged as a powerful framework, with approaches primarily distinguished by their treatment of state and action spaces \cite{Mnih2015,MJFHPV}. A prime example is autonomous robotic navigation, which demands algorithms that are computationally efficient, robust, and capable of online operation \cite{MJFLSTM}. While deep reinforcement learning has demonstrated remarkable success across various domains, its practical deployment encounters several fundamental challenges that limit widespread adoption. A primary concern is the substantial computational requirements of deep neural networks, which often hinder online operation in resource-constrained environments. Additionally, these methods exhibit significant sensitivity to hyperparameter tuning, necessitating extensive experimentation and domain expertise to achieve optimal performance \cite{MJFDIWO}. Perhaps most critically, the black-box nature of deep neural networks results in limited interpretability, thereby restricting their application in safety-critical domains where decision transparency is essential. These limitations collectively motivate the exploration of alternative approaches that can maintain learning capability while offering greater practicality for real-world systems \cite{ORL}.

Building on the RL foundation, value-based methods have demonstrated remarkable effectiveness in discrete action spaces. The Deep Q-Network (DQN)~\cite{Mnih2015} pioneered deep RL by combining Q-learning with neural network function approximation, while Double DQN~\cite{VanHasselt2016} mitigated overestimation bias through decoupled action selection and evaluation. Dueling DQN~\cite{Wang2016} further improved learning efficiency by separating value and advantage streams, and Rainbow DQN~\cite{Hessel2018} integrated multiple advancements-including prioritized experience replay and distributional RL-into a unified architecture with superior performance.

In contrast, for continuous control tasks, actor-critic methods have become the dominant paradigm. Deep Deterministic Policy Gradient (DDPG)~\cite{DDPG2024} extended DQN to continuous actions using deterministic policies, while Twin Delayed DDPG (TD3)~\cite{Fujimoto2018} improved training stability through twin critics and target policy smoothing. More recently, Soft Actor-Critic (SAC)~\cite{Haarnoja2018} introduced entropy regularization to encourage exploration and improve robustness, making it a strong baseline for complex, continuous domains.

Despite these advances, deep RL methods continue to face fundamental challenges. They typically require extensive hyperparameter tuning and considerable computational resources \cite{MJFSSRN}. Performance is highly sensitive to neural architecture design-such as the number of layers, neurons, and activation functions-yet principled guidelines for selecting these configurations remain elusive \cite{Zhang2022Overestimation}. Furthermore, policies represented by deep neural networks often exhibit limited robustness to input noise and environmental variations, and their black-box nature hinders interpretability and formal safety verification in critical applications \cite{SRL2025}.

These challenges have motivated alternative paradigms that preserve learning capability while improving interpretability and implementation simplicity. Fuzzy systems offer a structured rule-based representation with smooth interpolation and relatively compact parameterization \cite{Juang2024, Chen2024}. However, existing fuzzy RL approaches still face limitations in scalability and learning efficiency. To address these limitations while preserving interpretability, the Fuzzy Q-Learning framework introduced in \cite{fuzzy_Q} provides a baseline for integrating fuzzy inference with Bellman equilibrium. While effective in moderate-scale problems, this approach faces scalability limitations as the state--action dimension and partition granularity increase, leading to slower convergence and higher computational cost \cite{fuzzyQ2025, supervisedRL}.

In this work, we enhance the Fuzzy Q-Learning structure with a fuzzified experience replay mechanism and segmented eligibility traces. This integration enables multi-step credit assignment and improved reuse of past interactions within a continuous state--action formulation. The resulting Enhanced Fuzzy Q($\lambda$)-learning approach is aimed at improving stability and sample efficiency while preserving interpretability in moderate-scale continuous-control settings.

The key contributions of this paper are summarized as: \textbf{(I)} We integrate the Fuzzified Bellman Equation (FBE) with fuzzy eligibility traces and segmented experience replay, enabling multi-step credit assignment within a continuous fuzzy state--action representation. \textbf{(II)} We formulate an interpretable rule-based alternative to neural function approximation for moderate-scale continuous-control problems. \textbf{(III)} We provide a contraction-based analysis of the proposed fuzzified Bellman operator, establishing convergence of the learning process and achieving the suboptimal policy. \textbf{(IV)} We validate the method on the Cart--Pole benchmark against $n$-step fuzzy Q-learning, fuzzy SARSA($\lambda$), and a DDPG baseline.

\section{FUNDAMENTAL Q-LEARNING}
\noindent
The fundamentals of standard Q-learning and Fuzzy-QL are explained, here, highlighting their limitations when applied to complex continuous state-action spaces. In a discrete state space $s \in \mathcal{S}$ and a discrete action space $a \in \mathcal{A}$, in Q-learning, based on the Bellman optimality equation, the error of action-value function ($Q(s,a)$) estimation, as a temporal difference (TD), $\delta_t$, can be computed. Considering $\gamma \in (0,1)$ as a discount factor and $r_t$ as the reward received at step $t$:
\begin{equation}
 \delta_t = r_t + \gamma \max_{a'} Q(s_{t+1},a') - Q(s_t,a_t),
 \label{eq:delta}
\end{equation}
Then $Q(s,a)$ will be updated by learning rate $\alpha > 0$: 
\begin{equation}
 Q(s_t,a_t) \leftarrow Q(s_t,a_t) + \alpha \delta_t,
 \label{eq:q-learning}
\end{equation}
While this method is simple and convergent in tabular settings, it is limited to discrete spaces and one-step temporal credit assignment. To address this gap, in \cite{fuzzy_Q}, Gaussian membership functions are employed to partition continuous state and action spaces:
\begin{align}
\mu_{S_i}(s) &= \exp\left(-\frac{(s-c_i^S)^2}{2\sigma_S^2}\right), \quad i=1,\ldots,N_s \label{eq:state_mf} \\
\mu_{A_j}(a) &= \exp\left(-\frac{(a-c_j^A)^2}{2\sigma_A^2}\right), \quad j=1,\ldots,N_a \label{eq:action_mf}
\end{align}
Now, the Fuzzy Q-table, $\widehat{Q}$, contains entries $\widehat{Q}_{i,j}$ associated with each pair of fuzzy membership functions $(\mu_{S_i}(s),\mu_{A_j}(a))$. 

Normalized membership weights, which form a fuzzy distribution over state sets that ensure  $\Sigma_{i=1}^{N_s} w_i(s)=1$, are computed as:
\begin{equation}
w_i(s) = \frac{\mu_{S_i}(s)}{\sum_{k=1}^{N_s} \mu_{S_k}(s)} \label{eq:normalized_weights}
\end{equation}

The Fuzzified-TD error is then:
\begin{equation}
	\delta_{i,j}(t) = r_t + \gamma \Upsilon(s_{t+1}) - \widehat{Q}_{i,j}(t) \label{eq:td_error}
\end{equation}

where the fuzzified next-state value estimation is:
\begin{equation}
	\Upsilon(s') = \sum_{i=1}^{N_s} w_i(s')\max_j \widehat{Q}_{i,j} \label{eq:fuzzy_value}
\end{equation}
While the FQL-FBE method solved the issue with continuous state-action space, the learning process is long and it is not sample efficient.

\section{PROPOSED METHOD: Enhanced-FQL($\lambda$)}
In this paper we enhanced the Fuzzy--QL with fuzzified eligibility traces and experience replay, to introduce Enhanced-FQL($\lambda$). 
\subsection{Multi-Step Learning with Fuzzified Eligibility Traces}
We extend the fuzzy Q-learning with fuzzified eligibility traces for multi-step credit assignment. A Fuzzified Activation Matrix, $\zeta(s,a)$, is introduced as the observed experience. Then, fuzzified eligibility matrix, $0\leq E(t)\leq 1$, is detailed. 
\begin{equation}
	\zeta(s,a) = \mu_S(s) \mu_A(a)^\top \label{eq:activation_matrix}
\end{equation}
\begin{equation}
E_{i,j}(t) = \min \{ \gamma\lambda E_{i,j}(t-1) + \zeta_{i,j}(s_t,a_t) , 1\} \label{eq:eligibility_update}
\end{equation}
where $\lambda \in [0,1]$ is the trace-decay parameter. In addition to increasing learning efficiency, this new fuzzified eligibility trace method still maps experiences from their continuous space to a discrete space and re-expresses them in a discrete tabular representation. On the one hand, it does not have the complexity of experiences in continuous space, and on the other hand, it does not have the discrete space limitation of the tabular method in TD($\lambda$). This makes this method suitable for continuous state-action spaces or complex high-dimensional dynamics. Then, the Fuzzy Q-table Bellman update rule combines fuzzified temporal difference with fuzzified eligibility traces:
\begin{equation}
\widehat{Q}_{i,j}(t+1) = \widehat{Q}_{i,j}(t) + \alpha E_{i,j}(t) \delta_{i,j}(t) \label{eq:q_update}
\end{equation}

\subsection{Sample-Efficient Segmented Experience Replay}
To enable experience replay while preserving temporal dependencies for eligibility traces, we employ a segment-based replay buffer $\mathcal{D}$:
\begin{equation}
\mathcal{D} = \left\{\mathcal{S}_L^{(k)}\right\}_{k=1}^K, \quad \mathcal{S}_L = \left\{(s_\ell, a_\ell, r_\ell, s_{\ell+1})\right\}_{\ell=1}^L \label{eq:segment_buffer}
\end{equation}
where $\mathcal{D}$ denotes the replay buffer storing $K$ segments, each segment $\mathcal{S}_L$ is a sequence of $L$ transitions $(s_\ell, a_\ell, r_\ell, s_{\ell+1})$, with $s_\ell$, $a_\ell$, $r_\ell$, and $s_{\ell+1}$ representing the state, action, reward, and next state at step $\ell$, respectively. Here, a \textit{segment} refers to a contiguous sequence of state-action-reward transitions $(s_\ell, a_\ell, r_\ell, s_{\ell+1})$ of fixed length $L$. 

To maintain temporal consistency in multi-step learning, eligibility traces are reconstructed for each sampled segment. This reconstruction ensures proper credit assignment across consecutive time steps within the segment, which is crucial for the stability and accuracy of the Fuzzy Q($\lambda$)-learning algorithm. This Trace Reconstruction Mechanism is detailed in Algorithm \ref{Alg:ER}.
\begin{algorithm}[H]
\caption{Segment-Based Experience Replay with Trace Reconstruction}
\begin{algorithmic}[1]
\State \textbf{Input:} Replay buffer $\mathcal{D}$, segment length $L$, batch size $B$, $\gamma$, $\lambda$
\State Sample $B$ segments $\{\mathcal{S}_L^{(b)}\}_{b=1}^B$ from $\mathcal{D}$
\For{each segment $\mathcal{S}_L^{(b)}$}
    \State Initialize $E^{(b)} \leftarrow \mathbf{0}$
    \For{$\ell = 1$ to $L$}
        \State Extract $(s_\ell, a_\ell, r_\ell, s_{\ell+1})$
        \State Compute $\zeta(s_\ell, a_\ell)$ via \eqref{eq:activation_matrix}
        \State Update traces: $E^{(b)}$ via \eqref{eq:eligibility_update}
        \State Compute TD error: $\delta_{i,j}$ via \eqref{eq:td_error}
    \EndFor
    \State Batch update: $\widehat{Q} \leftarrow \widehat{Q} + \frac{\alpha}{B}\sum_{b=1}^B \sum_{\ell=1}^L (\delta_{i,j}) E^{(b)}$
\EndFor
\end{algorithmic}\label{Alg:ER}
\end{algorithm}

Experience replay improves sample efficiency by allowing multiple learning updates from each collected experience and decorrelating data samples, resulting in faster and more stable learning with fewer environment interactions.

\subsection{Action Selection and Exploration}
The greedy action is computed through a two-stage process:
\begin{equation}
	a^*(s) = \sum_{i=1}^{N_s} p_i(s) \cdot c_{j^*(i)}^A \label{eq:defuzzified_action}
\end{equation}
with the policy distribution:
\begin{equation}
	p_i(s) = \frac{\exp\left(w_i(s) \max_j \widehat{Q}_{i,j} / \beta\right)}{\sum_{k=1}^{N_s} \exp\left(w_k(s) \max_j \widehat{Q}_{k,j} / \beta\right)} \label{eq:policy_distribution}
\end{equation}
and greedy action indices:
\begin{equation}
	j^*(i) = \arg\max_j \widehat{Q}_{i,j} \label{eq:greedy_index}
\end{equation}

An $\epsilon$-greedy exploration strategy is employed during training to balance exploration and exploitation.

\subsection{Integrated Learning Algorithm}
The proposed Enhanced-FQL($\lambda$) algorithm is outlined in Algorithm \ref{Alg:Proposed}.
\begin{algorithm}
\caption{Fuzzy Q($\lambda$)-Learning with Experience Replay (Enhanced-FQL($\lambda$))}
\begin{algorithmic}[1]
\State \textbf{Input:} Fuzzy partitions, $\alpha$, $\gamma$, $\lambda$, $\epsilon$, $\beta$, $B$, $L$
\State Initialize $\widehat{Q}$, $E \gets 0$, $\mathcal{D} \gets \emptyset$
\State Observe $s_0$
\For{$t = 0, 1, 2, \ldots$}
    \State Compute $\mu_S(s_t)$ via \eqref{eq:state_mf}
    \State Compute $w_i(s_t)$ via \eqref{eq:normalized_weights}
    \State Compute $a_t$ via $\epsilon$-greedy policy via \eqref{eq:defuzzified_action}
    \State Execute $a_t$; observe $r_t$, $s_{t+1}$
    \State Store transition in current segment
    \If{segment full}
        \State Store segment in $\mathcal{D}$
    \EndIf
    \State Compute $\Upsilon(s_{t+1})$ via \eqref{eq:fuzzy_value}
    \State Update $E$ via \eqref{eq:eligibility_update}
    \State Compute $\delta_{i,j}$ via \eqref{eq:td_error}
    \State Update $\widehat{Q}$ via \eqref{eq:q_update}
    \If{replay condition}
        \State Execute Algorithm 1
    \EndIf
    \State $s_t \gets s_{t+1}$
\EndFor
\end{algorithmic}\label{Alg:Proposed}
\end{algorithm}

\section{Enhanced-FQL($\lambda$) Convergence Study}

The convergence and stability of the proposed Enhanced-FQL($\lambda$) method are established under the following standard assumptions.

\paragraph*{Assumption 1} The reward function is uniformly bounded, i.e., $|r_t| \le r_{\max} < \infty$. The fuzzy membership functions are bounded within $[0,1]$, and the normalized activation weights $0\leq w_i(\cdot)\leq 1$ form a convex combination, satisfying $\sum_i w_i(\cdot) = 1$.

\paragraph*{Assumption  2} The environment is modeled as a stationary Markov Decision Process with finite or compact state and action spaces. Under a behavior policy with sufficient exploration (e.g., $\epsilon$-greedy with decaying $\epsilon$), the induced state-action Markov chain is ergodic and admits a unique stationary distribution. Furthermore, the fuzzy rule base experiences persistent excitation, ensuring that all rule pairs $(i,j)$ are sufficiently activated.

\paragraph*{Assumption 3} Learning rates ($\alpha_t$) satisfy the Robbins–Monro conditions ($\sum_{t=0}^{\infty} \alpha_t = \infty$ and $\sum_{t=0}^{\infty} \alpha_t^2 < \infty$), guaranteeing both sufficient exploration and eventual convergence.

\paragraph*{Assumption 4} The discount factor $\gamma$ and trace-decay parameter $\lambda$ lie within $[0,1]$.

\paragraph*{Assumption 5} When computing the greedy action via defuzzification, the weight calculation (e.g., using a SoftMax over the per-rule maximum Q-values) ensures non-negative, normalized weights, thereby guaranteeing bounded control outputs and avoiding destabilizing negative weight effects.

\paragraph*{Assumption 6} To ensure stability under off-policy learning, Watkins's Q($\lambda$) is used with trace reset upon exploratory actions.

\begin{definition}
For any table $Q$, we define a fuzzified optimality operator.
\begin{equation}
	(\mathcal T_F Q)_{i,j} \; \triangleq \; \mathbb E\!\big[r_t\mid(i,j)\big] \;+\; \gamma\,\mathbb E\!\big[\Upsilon_Q(s_{t+1})\mid(i,j)\big],
\end{equation}
\end{definition}

\paragraph*{Boundedness remarks.}
(i) Since $E_{i,j}(t)\in[0,1]$, the per-step increment $\alpha_t E_{i,j}\delta_{i,j}$ is bounded for bounded $\widehat Q$ and rewards. (ii) The greedy action constructed as a convex combination of fixed action centers remains bounded, yielding bounded control inputs.

\subsection{Convergence Analysis}

\begin{theorem}
	\label{thm:convergence}
	Under Assumptions A1--A6, the sequence $\{\widehat{\mathbf{Q}}_t\}$ generated by the Enhanced-FQL($\lambda$) algorithm converges to a fixed suboptimal point $\widehat{\mathbf{Q}}^\star$ of the fuzzified Bellman optimality operator $\mathcal{T}_F$, i.e., $\lim_{t \to \infty} \widehat{\mathbf{Q}}_t = \widehat{\mathbf{Q}}^\star$.
	Consequently, the induced greedy-defuzzified policy converges to the fuzzy-optimal policy over the chosen rule base.
\end{theorem}

\begin{proof}
	The fuzzified Bellman operator $\mathcal{T}_F$ is a contraction mapping. In other words, for any two fuzzy Q-functions $\widehat{\mathbf{Q}}_1$ and $\widehat{\mathbf{Q}}_2$, we have:
	
	\[
	\begin{aligned}
		&\| \mathcal{T}_F \widehat{\mathbf{Q}}_1 - \mathcal{T}_F \widehat{\mathbf{Q}}_2 \|_\infty \\
		&= \max_{i,j} \left| \mathbb{E} \left[ r + \gamma \Upsilon_1(s') - r - \gamma \Upsilon_2(s') \right] \right| \\
		&= \gamma \max_{i,j} \left| \mathbb{E} \left[ \Upsilon_1(s') - \Upsilon_2(s') \right] \right| \\
		&\leq \gamma \max_{i,j} \mathbb{E} \left[ \left| \Upsilon_1(s') - \Upsilon_2(s') \right| \right] \\
		&\leq \gamma \max_{i,j} \mathbb{E} \left[ \sum_{k=1}^{N_s} w_k(s') \left| \max_l \widehat{Q}_{1,k,l} - \max_l \widehat{Q}_{2,k,l} \right| \right] \\
		&\leq \gamma \max_{i,j} \mathbb{E} \left[ \sum_{k=1}^{N_s} w_k(s') \max_m \left| \widehat{Q}_{1,k,m} - \widehat{Q}_{2,k,m} \right| \right] \\
		&\leq \gamma \max_{i,j} \mathbb{E} \left[ \max_{k,m} \left| \widehat{Q}_{1,k,m} - \widehat{Q}_{2,k,m} \right| \right] \\
		&= \gamma \| \widehat{\mathbf{Q}}_1 - \widehat{\mathbf{Q}}_2 \|_\infty.
	\end{aligned}
	\]
	
	Since $\gamma \in (0,1)$ by A4, $\mathcal{T}_F$ is a $\gamma$-contraction in the sup-norm. By the Banach fixed-point theorem, there exists a fixed suboptimal point $\widehat{\mathbf{Q}}^\star$ such that $\mathcal{T}_F \widehat{\mathbf{Q}}^\star = \widehat{\mathbf{Q}}^\star$.
	
\end{proof}

\begin{remark}
The fuzzy-optimal policy $\pi^\star$ is defined through the defuzzified greedy action in Eq. (12)-(14). The convergence of $\widehat{\mathbf{Q}}_t$ to $\widehat{\mathbf{Q}}^\star$ implies the convergence of $\pi_t$ to $\pi^\star$ due to the Lipschitz continuity of the SoftMax-based action selection mechanism.
\end{remark}
	
\section{SIMULATION RESULTS}
The proposed Enhanced-FQL($\lambda$) algorithm was evaluated on the classical cart and pole benchmark, a standard continuous control task that presents fundamental challenges in nonlinear stabilization. The environment simulates a pendulum with continuous state and action spaces, where the objective is to swing up and balance the pendulum in the upright position while minimizing control effort. The state space is defined by $s_t = [x, \dot{x}, \theta, \dot{\theta}]^\top$, where $\theta$ represents the pole angle, $\dot{\theta}$ the angular velocity, $x$ the cart position, and $\dot{x}$ the cart linear speed. The action space consists of continuous force values to the cart $a_t \in [-2, 2]$ N. The reward function is formulated to encourage stabilization with minimal energy consumption:
\[
r_t = -(\theta^2 + 0.1\dot{\theta}^2 + 0.001x^2 + 0.0001\dot{x}^2 +0.001a_t^2),
\]
penalizing state deviation and large control inputs.

For fuzzy approximation, Gaussian membership functions were employed to partition the state space into $N_s = \{3,3,7,5\} = 315$ fuzzy sets and the action space into $N_a = 5$ sets. The learning parameters were tuned experimentally, with learning rate $\alpha = 0.005$, discount factor $\gamma = 0.99$, trace decay $\lambda = 0.8$, segment length $L = 10$, batch size $B = 32$, and exploration rate $\varepsilon$ decaying linearly from $0.2$ to $0.05$ over the first 500 episodes. The segment length $L=10$ was selected as a practical compromise between preserving short-horizon temporal consistency for eligibility-trace reconstruction and maintaining sufficient decorrelation in replay; shorter segments truncate the effective trace more aggressively, whereas substantially longer segments reduce the replay benefit.

To evaluate the proposed Enhanced-FQL($\lambda$) method, a comparative study is conducted including: (i) a multi-step fuzzy Q-learning variant with $n$-step returns within the same fuzzy rule base, (ii) fuzzy SARSA($\lambda$) as an on-policy baseline with eligibility traces to assess the effect of the off-policy optimality operator, and (iii) DDPG using deep neural networks. For the DDPG baseline, the actor was implemented as a fully connected neural network with two hidden layers of 128 units each and ReLU activations, followed by a $\tanh$ output layer and a scaling layer mapping the action to the admissible force range. The critic was implemented as a state--action value network with separate state and action inputs, followed by two hidden layers of 256 units each with ReLU activations and a scalar output. Experience replay was used with a buffer capacity of $6\times10^5$ transitions and a minibatch size of 256, with parameter updates starting after 5000 stored transitions. Target networks were updated using soft updates with coefficient $\tau=5\times10^{-3}$. Exploration was induced by additive Gaussian action noise with initial standard deviation $1.5$, decayed multiplicatively by $0.997$, and actions were clipped to the admissible bounds. The discount factor was set to $\gamma=0.99$ and all results were averaged over 5 random seeds. This setup enables a clear analysis of the contribution of each algorithmic component to overall performance.

\begin{figure}[h]
	\centering
	\includegraphics[width=1\linewidth]{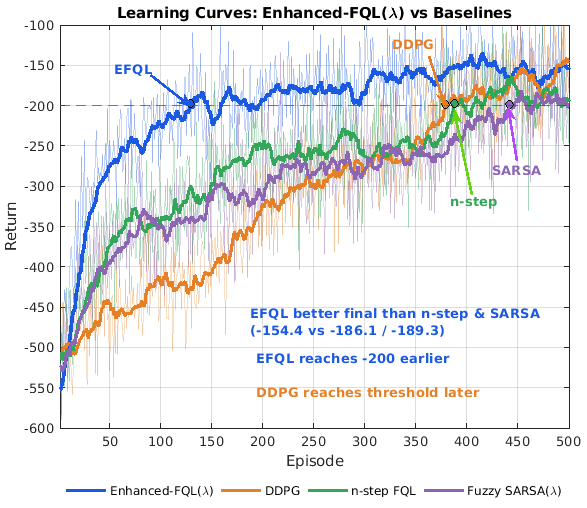}
	\caption{Learning curves comparing Enhanced-FQL($\lambda$) with baseline methods on Cart\&Pole benchmark. The proposed method demonstrates faster convergence and competitive performance.}
	\label{fig:pendulum_results}
\end{figure}

Figure~\ref{fig:pendulum_results} illustrates the learning progression over $500$ training episodes. Enhanced-FQL($\lambda$) reached the target return threshold in approximately $129$ episodes, converging faster than the fuzzy baselines and remaining competitive with the tested DDPG configuration on this benchmark. All methods were trained under the same environment, reward design, action bounds, episode horizon, random-start protocol, seed set, and evaluation procedure, while each baseline used its own algorithm-specific architecture and hyperparameterization.

Table~\ref{tab:results} provides a quantitative comparison across multiple performance metrics. Enhanced-FQL($\lambda$) achieved the best average return among the evaluated methods and reached the target threshold earlier than the fuzzy baselines. The tested DDPG baseline delivered competitive final performance, but with higher per-update computation in the present implementation.

\begin{table}[h]
\centering
\caption{Performance comparison on Cart\&Pole environment}
\begin{tabular}{lccc}
\hline
\textbf{Algorithms} & \textbf{a} & \textbf{b} & \textbf{c} \\
\hline
n-step FQL & $-186$ & $0.42$ & $388$ \\
Fuzzy SARSA($\lambda$) & $-197$ & $0.45$ & $442$ \\
DDPG & $-166$ & $0.80$ & $379$ \\
\textbf{Enhanced-FQL($\lambda$)} & $\mathbf{-159}$ & $\mathbf{0.48}$ & $\mathbf{129}$ \\
\hline
\end{tabular}\\
\label{tab:results}
\begin{flushleft}\footnotesize{a: Average Return in the last 10\% episodes; b: Update Time (ms); Convergence episode}\end{flushleft}
\end{table}

The comparison study supports the effectiveness of integrating fuzzy eligibility traces and segment-based experience replay into FQL--FBE. Three observations are particularly relevant.

First, the multi-step credit assignment provided by fuzzy eligibility traces ($\lambda = 0.8$) accelerated learning relative to the one-step fuzzy baseline, reducing the sample requirement for convergence by approximately 35\% compared to $n$-step FQL.

Second, the segment-based experience replay mechanism effectively decorrelated training samples while maintaining temporal consistency, resulting in the lowest variance. This stability is particularly valuable where training data may be limited or expensive to acquire.

Third, the comparison with DDPG indicates that the proposed method can remain competitive on this benchmark while preserving an interpretable rule-based representation.

The fuzzy rule base also permits direct inspection of the learned control structure, which is a practical advantage when interpretability is important.

\begin{figure}
	\includegraphics[width=1\linewidth]{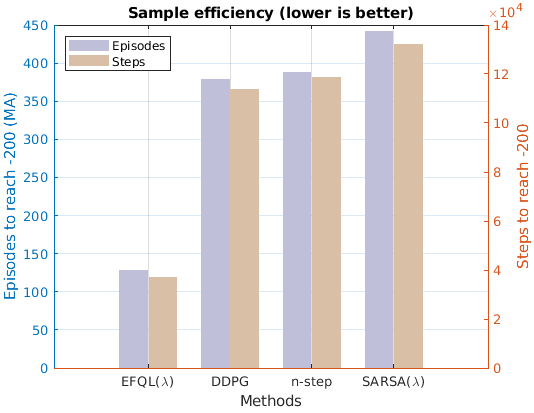}
	\caption{Sample efficiency comparison}
	\label{fig:numerical_results}
\end{figure}

Figure~\ref{fig:numerical_results} compares the speed of reaching a target performance threshold (return = -200). Enhanced-FQL($\lambda$) reaches this threshold in fewer episodes than the fuzzy baselines and earlier than the tested DDPG configuration on this benchmark.

\section{DISCUSSION}
The computed value function \(\widehat{Q}^\star\) represents the suboptimal fixed point within the chosen fuzzy rule base. While the proposed method involves an inherent bias-approximation trade-off; refining the state and action partitions---that is, increasing \(N_s\) and \(N_a\)---can reduce approximation bias, it also leads to increased computational complexity. Furthermore, to ensure stability in off-policy control settings, our implementation adopts the Watkins Q(\(\lambda\)) approach, wherein eligibility traces are reset whenever the executed action diverges from the greedy target policy. This mechanism preserves the contraction property of the Bellman optimality operator and facilitates convergent learning behavior.

The bias--variance trade-off and its impact on sample efficiency are fundamental considerations in temporal-difference learning. In $n$-step TD methods, increasing $n$ reduces bias at the cost of higher variance, vice versa. TD($\lambda$) provides a continuous mechanism to balance this trade-off through the trace decay parameter $\lambda$. The proposed Enhanced-FQL($\lambda$) framework inherits this flexible trade-off and further reduces variance by enabling credit assignment across multiple active fuzzy rules via soft generalization. This attribute often leads to improved sample efficiency, particularly in continuous or noisy domains.

The behavior of Enhanced-FQL($\lambda$) can be analyzed through its limit cases with respect to the trace decay parameter $\lambda$. As $\lambda$ approaches zero, the algorithm reduces to one-step fuzzy Q-learning, effectively suppressing eligibility traces and recovering the base FQL-FBE update. Conversely, as $\lambda$ tends toward one, the method approaches Monte Carlo-style credit assignment within the fuzzy basis, propagating returns across entire episodes, subject to episodic finiteness. These limit cases illustrate the continuum of temporal credit assignment strategies encompassed by the proposed framework.

The comparative results reveal three key findings: First, in terms of sample efficiency, Enhanced-FQL($\lambda$) with moderate $\lambda$ values (e.g., 0.5--0.9) achieved target performance levels in fewer episodes compared to $n$-step TD, which can be attributed to continuous credit assignment across activated fuzzy rules. Second, regarding stability, the combination of SoftMax-weighted defuzzification and fuzzy Bellman backups resulted in lower variance in learning curves compared to tabular $n$-step methods, particularly under sensor noise conditions. Third, ablation studies confirmed our design choices by demonstrating performance degradation when using unnormalized value backups ($\Upsilon$), disabling traces ($\lambda=0$), and switching to on-policy SARSA($\lambda$)-style updates. 

\section{CONCLUSION}
This paper introduced Enhanced-FQL($\lambda$), a fuzzy reinforcement learning framework that combines fuzzified eligibility traces, segmented experience replay, and a fuzzified Bellman backup within an interpretable rule-based representation. Theoretical analysis established convergence under standard assumptions to the fixed point induced by the selected fuzzy basis. On the Cart--Pole benchmark, the proposed method improved sample efficiency and reduced variance relative to the fuzzy baselines, while remaining competitive with the tested DDPG configuration. These results support Enhanced-FQL($\lambda$) as a promising interpretable alternative for moderate-scale continuous-control problems, while broader validation on more complex benchmarks remains an important direction for future work.

\end{document}